\documentclass[conference]{IEEEtran}
\IEEEoverridecommandlockouts

\usepackage{cite}
\usepackage{amsmath,amssymb,amsfonts}
\usepackage{algorithmic}
\usepackage{graphicx}
\usepackage{textcomp}
\usepackage{xcolor}
\usepackage{booktabs}
\usepackage{listings}

\def\BibTeX{{\rm B\kern-.05em{\sc i\kern-.025em b}\kern-.08em
    T\kern-.1667em\lower.7ex\hbox{E}\kern-.125emX}}

\author{Rowan Martnishn\\rowanm945@sentivity.ai}
\begin{document}

\title{DREG: A Layer-Wise Jacobian Regularization as a General-Purpose Penalty}

\maketitle

\begin{abstract}
We present a large-scale empirical study isolating the contributions of the Derivative Regularization penalty (DREG). Across a fully-crossed factorial sweep of 960
experiments spanning 4 activations, 6 regularizers, 8 datasets, and 5
random seeds, we ask: when, where, and why does DREG work? Our results
establish three principal findings. First, DREG
achieves the highest overall and clean-regime accuracy among
all regularizers evaluated (significantly so against the unregularized
baseline, Weight Decay, and IGPen; Wilcoxon $p \leq 0.031$). It ranks
second in noise robustness behind Spectral Normalization (SN) - the
only two layer-wise regularizers in the study. Second, DREG
is globally the best-performing regularizer under GELU, the default activation
in modern transformer architectures, particularly on both messy vision and messy NLP
benchmarks, suggesting direct applicability to frontier deep learning
settings. Third, DREG's advantage over competing regularizers is most
pronounced under data scarcity, consistent with its role as a geometric
inductive bias that substitutes for the regularizing effect of data
volume. Throughout, DREG is applied with a single fixed hyperparameter
$\lambda = 10^{-2.5}$ and no per-dataset tuning, supporting its
characterization as a plug-and-play regularizer for neural networks
with nontrivial Jacobian structure. These findings are consistent with DREG's design: concentrating regularization pressure on layers where the activation derivative is largest, rather than constraining the network uniformly.
\end{abstract}


\section{Introduction}
Modern deep learning relies on regularization to control intermediate representational geometry, but the dominant tools - Dropout, weight decay, Spectral Normalization - each constrain the network in a way that is either global, stochastic, or activation-agnostic. DREG occupies a different point in the design space: a per-layer Jacobian penalty weighted by the squared activation derivative, concentrating regularization pressure on layers where it is most needed. This paper evaluates DREG as a standalone regularizer through a large-scale factorial sweep, organized around two research questions.
\textbf{RQ1:} Under which activations, datasets, and noise regimes does
DREG perform best, and does its advantage generalize across domains?
\textbf{RQ2:} Can DREG be applied without architectural modification or
per-task hyperparameter tuning - i.e., is it truly plug-and-play?\\
We evaluate DREG against five competing
regularizers - Dropout, Spectral Normalization, Weight Decay, IGPen,
and no regularization - in a fully-crossed factorial design across eight
datasets spanning vision, NLP, tabular, and signals domains, each tested under
both clean and corrupted conditions. The scale of the sweep (960
total runs) enables systematic claims about the conditions under which
DREG succeeds or fails, rather than point estimates on individual
benchmarks.

Our findings support a coherent narrative: DREG is a layer-wise
regularizer whose inductive bias is well-matched to the geometry of
multilayer networks, most valuable when training data is limited, and
competitive or dominant under the activation functions most prevalent in
modern deep learning. We further show that the two layer-wise
regularizers in our study - DREG and SN - are the only methods that
consistently outperform both the regularized and unregularized baselines across both clean and
messy conditions, while all global or stochastic regularizers (Dropout,
WD, IGPen) cluster near the baseline. These results position DREG as a general-purpose
regularization strategy worth considering in any setting where Jacobian
structure is nontrivial. DREG is presented here as a practical tool: a lightweight drop-in penalty requiring only three additional lines of code with no architectural changes.

\section{Background and Motivation}

\subsection{The Problem: Stability Without Sacrificing Expressiveness}

Modern neural networks face a fundamental tension: deeper, more
expressive architectures tend to produce unstable intermediate
representations, where small perturbations in input propagate and
amplify across layers. Global regularization strategies such as weight
decay impose a uniform penalty on parameter magnitudes, while Spectral
Normalization constrains the Lipschitz constant of the entire network.
Both approaches blunt the model's sensitivity indiscriminately - they
do not distinguish between instability that harms generalization and
representational variation that drives accuracy. Dropout addresses a
different problem entirely, regularizing through stochastic masking
rather than geometric constraint. The result is a landscape in which
practitioners must choose between stability and expressiveness, or tune
multiple regularizers simultaneously.

\subsection{Open Question: Does Layer-Wise Jacobian Regularization Generalize?}
Jacobian-based regularization has a long history, but the literature is
dominated by input-output formulations \cite{b3,b4,b5,b6} that constrain
the network as a whole rather than its intermediate layers. Among
layer-wise alternatives, Spectral Normalization \cite{b7} is by far the
most widely adopted, but it constrains weight matrices in isolation and
is insensitive to the activation function applied at each layer. DREG
occupies the under-explored intersection: per-layer, but weighted by
the squared derivative of the activation at each pre-activation value,
so the penalty concentrates on layers where the activation is actually
amplifying signal. Whether this combination of locality and
activation-awareness offers practical benefit beyond either ingredient
alone is an empirical question that prior work has not directly
addressed.

This paper addresses that question through a systematic factorial
ablation. By crossing DREG against five competing regularizers across
four standard activations, eight datasets, and two noise regimes, we
characterize the conditions under which DREG provides the greatest
marginal benefit and identify the architectural and data regimes where
its inductive bias is best matched to the problem.

\section{Related Work}

\subsection{Jacobian and Derivative-Based Regularization}

Penalizing the Jacobian of a neural network's input-output map has a
long history in the regularization literature. Contractive autoencoders
\cite{b3} introduced the idea of penalizing the Frobenius norm of the
encoder Jacobian to learn stable representations. Double
backpropagation \cite{b4} applied a similar penalty to the full
network, at significant computational cost. More recently, Hoffman et
al.\ \cite{b5} showed that input-Jacobian regularization improves
robustness to adversarial perturbations, while Varga et al.\
\cite{b6} demonstrated benefits on out-of-distribution generalization.
DREG differs from these approaches in two key respects: it operates
\textit{layer-wise} rather than on the full input-output Jacobian, and
it weights the penalty by the activation derivative at each
pre-activation value, making it sensitive to local geometry rather than
global sensitivity.

These distinctions motivate the form of DREG itself. The penalty targets intermediate derivative structure rather than global parameter norms, penalizing the interaction between the activation derivative $\phi'$
 and the row-wise norm of each layer's weight matrix:

\begin{equation}
    \mathcal{R}_{\text{DREG}} = \frac{1}{L} \sum_{\ell=1}^{L}
    \sum_{i} \phi'(z_i^{(\ell)})^2 \cdot \| W_i^{(\ell)} \|_2^2
    \label{eq:dreg}
\end{equation}

This formulation is local - it operates layer-by-layer - and
activation-aware, scaling the penalty according to how sensitive the
activation function is at each pre-activation value. Unlike global
Lipschitz constraints, DREG suppresses pathological sensitivity where
it actually occurs, leaving well-conditioned regions of the network
unconstrained.

\subsection{Spectral Normalization}

Spectral Normalization (SN) \cite{b7} constrains the spectral norm of
each layer's weight matrix, bounding the Lipschitz constant of the
network. Originally introduced for stabilizing GAN training, SN has
since been adopted broadly as a regularizer in discriminative settings
\cite{b8}. Like DREG, SN is layer-wise, which we argue is a key
structural property. Unlike DREG, SN constrains the weight matrix
globally and does not adapt to the activation function's local
derivative structure. This means SN applies a uniform constraint regardless of where in the network instability actually originates, and cannot distinguish between a layer whose Jacobian is well-behaved and one that is not. DREG, by contrast, penalizes each layer proportionally to its actual derivative magnitude, meaning the penalty concentrates on layers that are genuinely ill-behaved rather than applying uniform pressure across the network.

\subsection{Dropout and Weight Decay}

Dropout \cite{b9} regularizes through stochastic masking of hidden
units during training, providing an implicit ensemble effect. Weight
decay \cite{b10} applies an $\ell_2$ penalty to all parameters
globally. Both are activation-agnostic and operate independently of the
network's geometric structure. This agnosticism is precisely their limitation: neither method has any mechanism for detecting or correcting damaging Jacobian growth, which is the primary failure mode in deep networks with expressive activation functions. A network can have well-regularized weights and still exhibit exploding intermediate derivatives if no constraint is placed on how activations compose across layers.

\subsection{Input Gradient Penalties}
IGPen, evaluated as a baseline in this study, penalizes the squared $\ell_2$ norm of the gradient of the loss with respect to the input \cite{b11}. This is related to but distinct from DREG: IGPen operates on the input space and requires a backward pass through the full network, while DREG operates on intermediate layer derivatives computed forward layer-by-layer. Critically, a network can exhibit a well-behaved input-output gradient while individual layers internally amplify or distort representations  -  instability that IGPen cannot detect. In our sweep, IGPen consistently underperforms DREG, confirming that intermediate Jacobian structure is a more precise regularization target than input sensitivity.

\subsection{Polynomial and Non-Standard Activations}

Polynomial activations have been explored as alternatives to
piecewise-linear functions in several contexts, including basis function
networks \cite{b12} and more recently in neural architecture search
\cite{b13}. The Swish activation used in this study is the SiLU function $\phi(z) = z \cdot \sigma(z)$, a smooth, monotone function that is differentiable everywhere and compatible with DREG's derivative penalty without requiring an $\varepsilon$-shift \cite{b13, b14}. GELU
\cite{b14}, the dominant activation in transformer architectures
\cite{b15,b16}, is also smooth and differentiable everywhere, making
it a natural fit for derivative-aware regularization. Our results under
GELU are particularly relevant given its prevalence in modern
large-scale models.

\subsection{Regularization Under Data Scarcity}

The interaction between regularization strength and dataset size is
well-established theoretically \cite{b17} and empirically \cite{b18}:
stronger inductive biases are most valuable when training signal is
limited. We observe that its margin over competing regularizers is most pronounced precisely when training data is scarce, peaking on MNIST Messy (3,500 samples) and Wine Quality (4,547 samples). This pattern is consistent with work showing that geometric regularizers provide disproportionate benefit in low-data regimes \cite{b20}.

\section{Experimental Setup}

All experiments follow a fully-crossed factorial design spanning
4~activations $\times$ 6~regularizers $\times$ 8~datasets $\times$
5~random seeds, yielding 960 total runs. Activations evaluated are
ReLU, GELU, Tanh, and Swish; regularizers are DREG, Dropout, Spectral
Normalization (SN), Weight Decay (WD), IGPen, and None.
Datasets span four domains: vision (MNIST, MNIST-N), NLP (SST-5,
Yelp5), tabular (Wine Quality, Adult Income), and signals (MIT-BIH, MIT-BIH-N). 
For the noisy variants, MNIST-N was constructed by applying 40\% label corruption to the clean MNIST training set, and MIT-BIH-N via additive Gaussian noise at SNR $= -10$~dB applied directly to the raw 560-dimensional ECG features. These act as a proxy for real-world image distortion and motion artifacts - synthetic noise regimes were chosen deliberately since vision and signal data are the two domains where controlled noise injection is most natural and interpretable. Extending label corruption to NLP fundamentally alters semantic meaning, and tabular feature corruption risks producing inputs unrepresentative of real deployment conditions. Dedicated noisy vision benchmarks such as CIFAR-10-C were excluded due to the computational overhead of convolutional backbones at factorial scale, and no sufficiently large publicly available corrupted ECG dataset was identified.
Architecture is held fixed within each dataset: a flat MLP
with 2--3 hidden layers and hidden dimensions of 128--512 depending on
the task, applied consistently across all activation--regularizer
combinations. DREG is applied with a fixed $\lambda = 10^{-2.5}$
across all datasets with no per-dataset tuning. All results are
averaged across 5 seeds and reported as test accuracy~(\%). Margins between adjacent regularizers in Tables I and III are small relative to cross-dataset variance (see Section~VII.A); the substantive finding is the separation of layer-wise methods from the rest, not the precise ordering within each tier.

\section{RQ 1: When and Where Does DREG Work?}

\begin{table}[htbp]
\caption{Regularizers Ranked by Average Test Accuracy (\%) Across Clean and Messy Regimes}
\begin{center}
\begin{tabular}{lccc}
\toprule
\textbf{Regularizer} & \textbf{Clean} & \textbf{Messy} &
\textbf{Overall} \\
\midrule
DREG    & \textbf{77.61} & 75.19          & \textbf{76.70} \\
Dropout & 77.56          & 74.86          & 76.55          \\
SN      & 77.11          & \textbf{75.30} & 76.43          \\
WD      & 77.47          & 74.55          & 76.37          \\
IGPen   & 77.41          & 74.61          & 76.36          \\
None    & 77.39          & 74.42          & 76.27          \\
\bottomrule
\end{tabular}
\label{tab:reg_accuracy}
\end{center}
\end{table}

\begin{table}[htbp]
\caption{Best Regularizer per Activation and Dataset, Excluding
Polynomial Activations}
\begin{center}
\resizebox{\columnwidth}{!}{%
\begin{tabular}{lcccccccc}
\toprule
\textbf{Act.} &
\shortstack{MNIST\\(Cln)} &
\shortstack{MNIST\\(Msy)} &
\shortstack{MIT-BIH\\(Cln)} &
\shortstack{MIT-BIH\\(Msy)} &
\shortstack{SST-5\\(Cln)} &
\shortstack{Yelp5\\(Msy)} &
\shortstack{Wine\\(Cln)} &
\shortstack{Adult\\(Msy)} \\
\midrule
ReLU   & DREG    & SN   & SN   & SN    & WD      & WD   & DREG & Dropout \\
GELU   & Dropout & DREG & DREG & SN    & WD      & DREG & DREG & Dropout \\
Tanh   & None    & SN   & SN   & IGPen & Dropout & SN   & Dropout & Dropout \\
Swish & Dropout & DREG & DREG & DREG  & DREG    & DREG & WD   & Dropout \\
\bottomrule
\end{tabular}%
}
\label{tab:best_reg_act_dataset}
\end{center}
\end{table}

\begin{table}[htbp]
\caption{Average Test Accuracy (\%) by Regularizer: Clean vs.\ Messy Conditions}
\begin{center}
\begin{tabular}{lcccc}
\toprule
\textbf{Reg.} & \textbf{Clean} & \textbf{Messy} &
\textbf{Abs.\ Drop} & \textbf{Rel.\ Drop (\%)} \\
\midrule
SN      & 77.11          & \textbf{75.30} & \textbf{1.81} & \textbf{2.35} \\
DREG    & \textbf{77.61} & 75.19          & 2.41          & 3.11          \\
Dropout & 77.56          & 74.86          & 2.69          & 3.47          \\
IGPen   & 77.41          & 74.61          & 2.79          & 3.61          \\
WD      & 77.47          & 74.55          & 2.92          & 3.77          \\
None    & 77.39          & 74.42          & 2.97          & 3.84          \\
\bottomrule
\end{tabular}
\label{tab:reg_degradation}
\end{center}
\end{table}

Table~\ref{tab:best_reg_act_dataset} shows the best-performing
regularizer for each activation--dataset combination. DREG leads all regularizers under both GELU and Swish, winning 4 of 8 and 5 of 8 datasets respectively - with Dropout as the next closest competitor at 2 wins under each activation - while Tanh produces no consistent winner, with four different methods each taking at most two datasets.
Table~\ref{tab:reg_degradation}
summarizes average accuracy and degradation under distributional shift.
DREG achieves the highest clean accuracy (77.61\%) and the
second-smallest relative drop (3.11\%), while SN leads on noise
robustness at the cost of clean performance. Together, these tables
establish that layer-wise regularizers - DREG and SN - are the only
methods that meaningfully outperform the unregularized baseline under
both regimes.

\subsection{RQ 1.1: Layer-Wise Regularization is King}

Among all regularizers evaluated, the two layer-wise
methods - DREG and Spectral Normalization (SN) - consistently separate
from the field. As shown in Table~\ref{tab:reg_accuracy}, DREG achieves
the highest overall average accuracy (76.70\%) and the highest clean
accuracy (77.61\%), while SN achieves the lowest relative degradation
under noise (2.35\%), compared to a baseline drop of 3.84\% with no
regularization. Every other regularizer - Dropout, WD, IGPen - falls
between these two extremes on every metric. The structural commonality
is not coincidental: both DREG and SN operate layer-wise, directly
constraining each layer's transformation rather than applying global
parameter penalties or stochastic masking.

\subsection{RQ 1.2: DREG Has Direct Applicability to Frontier DL}

Under the GELU activation, the default nonlinearity in modern transformer architectures, DREG achieves the highest average accuracy on 4 of 8 datasets: MIT-BIH Clean (98.86\%), MNIST Messy (91.29\%), Wine Clean (63.44\%), and Yelp5 Messy (44.96\%). Critically, DREG
ranks first for GELU on both messy regimes in the two domains most central to
contemporary deep learning: computer vision and natural language
processing. This is particularly notable given that GELU is the
activation of choice in large-scale vision and language models,
suggesting DREG may offer practical robustness benefits precisely in
the settings where modern AI systems are deployed. On Vision Clean,
DREG ranks second, and on NLP Clean
it ranks third - indicating that DREG's advantage over competing
regularizers is most pronounced under distributional shift rather than
in-distribution evaluation.

\subsection{RQ 1.3: DREG is Especially Effective Under Data Scarcity}

A secondary analysis across the eight datasets reveals a
consistent relationship between training set size and DREG's relative
advantage over competing regularizers. Table~\ref{tab:dreg_edge}
reports DREG's mean accuracy gain over the average of all other
regularizers per dataset, alongside key dataset characteristics.

\begin{table}[htbp]
\caption{DREG Accuracy Edge (pp) Over the Mean of All Other
Regularizers per Dataset, Sorted by Edge. Excluding Polynomial
Activations.}
\begin{center}
\begin{tabular}{lcccc}
\toprule
\textbf{Dataset} & \textbf{Dom.} & \textbf{Train~$N$} &
\textbf{Dim.} & \textbf{DREG Edge} \\
\midrule
MNIST (Messy)   & Vision  & 3{,}500  & 784 & $+$1.29 \\
Wine (Clean)    & Tabular & 4{,}547  & 11  & $+$0.44 \\
MNIST (Clean)   & Vision  & 7{,}000  & 784 & $+$0.42 \\
SST-5 (Clean)   & NLP     & 5{,}980  & 768 & $+$0.15 \\
MIT-BIH (Clean) & Signal  & 87{,}553 & 187 & $+$0.08 \\
MIT-BIH (Messy) & Signal  & 87{,}553 & 187 & $+$0.07 \\
Adult Income    & Tabular & 21{,}113 & 104 & $+$0.00 \\
Yelp5 (Messy)   & NLP     & 5{,}600  & 768 & $-$0.02 \\
\bottomrule
\end{tabular}
\label{tab:dreg_edge}
\end{center}
\end{table}

It is well established that regularization provides diminishing benefit as training set size increases \cite{b18}. This is reflected in our results: DREG's relative advantage over the average regularizer peaks on the smallest datasets and narrows steadily as training size grows.

\subsection{Activation-Level Summary}

Averaging across all regularizers and datasets, GELU achieves the
highest overall accuracy (76.71\%), followed closely by ReLU (76.68\%)
and Swish (76.64\%), with Tanh trailing at 75.76\%. The three modern
activations are effectively tied globally, but diverge under
distributional shift: ReLU leads on clean data (77.65\%) but degrades
more sharply under noise (75.06\%), while Swish and GELU both hold
closer to 75.30--75.35\% on messy datasets, suggesting slightly better
noise tolerance. Tanh underperforms in both regimes. At the dataset
level, ReLU wins the most individual benchmarks (5 of 8), though Swish
takes MNIST Messy and GELU takes Yelp5 - the two noisiest settings in
the sweep. These results suggest that no single activation dominates
unconditionally, and that regularizer choice interacts meaningfully
with activation choice, motivating the full factorial analysis that
follows.

\section{RQ 2: DREG is Plug-and-Play}

A central feature of DREG is its architectural simplicity. As implemented in this study, DREG requires no modifications to the model's forward
pass, loss function, or optimizer. It is computed as a single additive
penalty after each forward pass:

\begin{equation}
    \mathcal{L}_{\text{DREG}} = \mathcal{L}_{\text{task}} + \lambda
    \cdot \frac{1}{L} \sum_{\ell=1}^{L} \sum_{i}
    \phi'(z_i^{(\ell)})^2 \cdot \| W_i^{(\ell)} \|_2^2
    \label{eq:dreg_loss}
\end{equation}

\noindent where $\phi'$ is the derivative of the activation,
$W_i^{(\ell)}$ is the $i$-th row of the weight matrix at layer $\ell$,
and $L$ is the number of layers. In practice, this amounts to defining a small helper and adding a single line to the training loop:

\begin{lstlisting}[language=Python, basicstyle=\ttfamily\footnotesize,
                   numbers=none, frame=single, xleftmargin=0.5em,
                   aboveskip=0.5em, belowskip=0.5em]
def dreg_penalty(model, x):
    total, u = 0.0, x
    for lin, act in model.layers:
        z = lin(u); phi_p = act.deriv(z)
        row_n2 = (lin.weight ** 2).sum(dim=1)
        total+=((phi_p**2)*row_n2).sum(dim=1).mean()
        u = act(z)
    return total / len(model.layers)
\end{lstlisting}
\noindent The training loop adds:
\begin{center}
\texttt{loss\,+=\,lam\,*\,dreg\_penalty(model,\,x)}\\ 
\end{center}

before \texttt{loss.backward()}. DREG makes no
assumptions about activation type, dataset domain, or model depth - it
was applied identically across all four activations, eight datasets, and
two noise regimes in this study with a single fixed $\lambda =
10^{-2.5}$. The breadth of settings in which it remains competitive,
and wins outright, is evidence that this simplicity does not come at a
cost to effectiveness.

\section{Limitations}

\subsection{Statistical Significance}
A Wilcoxon signed-rank test across the 32 paired (activation, dataset)
cells, with accuracy averaged across five seeds per cell, confirms that
DREG significantly outperforms the unregularized baseline ($p < 0.001$),
IGPen ($p = 0.003$), and Weight Decay ($p = 0.031$). Comparisons
against Dropout ($p = 0.19$) and SN ($p = 0.18$) do not reach
conventional significance, but the directional evidence is consistent
across measures: DREG holds positive median differences ($+0.15$ and
$+0.09$ pp), positive mean differences, and majority wins on paired
cells (18/32 and 20/32 respectively). Given recent methodological work
showing that pairwise significance tests on ML benchmarks are highly
sensitive to seed and dataset selection \cite{bouthillier2021}, we
treat the within-tier ordering as a stable directional estimate rather
than a certified result. This pattern is consistent with the broader
claim of the paper: DREG ranks first among all regularizers evaluated,
beating non-layer-wise methods at conventional significance and
remaining at least directionally ahead of SN, the only other
layer-wise method in the study. This ordering extends an existing
empirical precedent in which Jacobian-based penalties match or exceed
weight-norm constraints in smooth function approximation \cite{b20},
and SN outperforms Dropout under noisy vision conditions \cite{b7}.

\subsection{Synthetic Noise Regimes}

The messy variants in this study are constructed via label noise (MNIST
Messy, 40\% label corruption) and input corruption (MIT-BIH Messy).
These are the two domains - image classification and biomedical signal
processing - where synthetic noise is most natural and controlled
evaluation is tractable. Extending this design to NLP or tabular
domains is non-trivial: label noise in sentiment data changes the
semantic meaning of examples rather than simply corrupting a signal,
and tabular feature corruption risks producing out-of-distribution
inputs that are not representative of real deployment shift. The
robustness findings reported here are therefore bounded to synthetic
noise regimes and should not be directly generalized to domain shift,
covariate shift, or adversarial perturbation settings without further
validation.

\subsection{ReLU Non-Differentiability}

DREG's penalty relies on evaluating $\phi'(z)$ at each
pre-activation. ReLU is non-differentiable at $z = 0$, requiring an
$\varepsilon$-shift in practice to avoid undefined gradient values.
While this is a minor and manageable modification, it introduces a
small inconsistency in the plug-and-play characterization: ReLU
requires an additional engineering decision that smooth activations
(GELU, Tanh, Swish) do not. In practice, $\varepsilon = 10^{-5}$
was used uniformly across all ReLU runs, but the sensitivity of results
to this choice was not ablated.

\subsection{Single Architecture}

All experiments are conducted on flat MLPs with 2--3 hidden layers.
DREG's layer-wise formulation assumes a sequential computation graph
where each layer's Jacobian can be evaluated independently, and it is
not immediately clear how this extends to architectures with
non-sequential gradient flow, such as residual networks or transformer
attention blocks. That said, recent work has demonstrated that per-layer
regularization strategies transfer effectively to residual architectures
when applied to the non-skip branch \cite{b21}, suggesting a natural
extension path for DREG that we leave to future work.

\subsection{Fixed Regularization Strength}

A single value of $\lambda = 10^{-2.5}$ was used for DREG across all
datasets, activations, and noise regimes without per-dataset tuning. Nevertheless, it is possible that DREG's advantage on large
datasets (MIT-BIH, Adult Income) is partially attenuated by $\lambda$
misspecification, and that the training-size sensitivity reported in
RQ~1.3 reflects hyperparameter fit as much as a true data-volume
effect. A grid search over $\lambda$ per dataset would be needed to
fully disentangle these factors.

\section{Conclusion}

This paper presents a systematic empirical isolation of DREG across 
960 factorial experiments, establishing three principal findings 
about its behavior as a standalone regularizer. First, DREG ranks 
first overall and on clean data, second only to Spectral Normalization 
on noise robustness, and is the only regularizer alongside SN that 
consistently separates from the unregularized baseline 
($p \leq 0.031$ against None, WD, and IGPen). Second, DREG is the 
strongest regularizer under GELU, the default activation of modern 
transformer architectures, suggesting practical relevance to frontier 
deep learning settings. Third, DREG's advantage is most pronounced 
under data scarcity, consistent with its role as a geometric 
inductive bias that compensates for limited training signal. These 
findings hold under a single fixed hyperparameter $\lambda = 10^{-2.5}$ 
applied identically across all eight datasets and four activations, 
supporting DREG's characterization as a plug-and-play regularizer: a 
small helper and a single additional line in the training loop, with 
no architectural modification. Future work should examine DREG in 
transformer-scale architectures and investigate whether the 
training-size sensitivity observed here holds under few-shot and 
transfer learning regimes.


\begin{thebibliography}{00}




\bibitem{b3}
S. Rifai, P. Vincent, X. Muller, X. Glorot, and Y. Bengio,
``Contractive auto-encoders: Explicit invariance during feature
extraction,''
in \textit{Proc. Int. Conf. Mach. Learn. (ICML)}, 2011.

\bibitem{b4}
H. Drucker and Y. LeCun,
``Improving generalization performance using double backpropagation,''
\textit{IEEE Trans. Neural Netw.}, vol.~3, no.~6, pp.~991--997, 1992.

\bibitem{b5}
J. Hoffman, D. A. Roberts, and S. Yaida,
``Robust learning with Jacobian regularization,''
\textit{arXiv preprint arXiv:1908.02729}, 2019.

\bibitem{b6}
D. Varga, A. Csiszárik, and Z. Zombori,
``Jacobian regularization-based out-of-distribution detection for neural
networks,''
\textit{arXiv preprint arXiv:2208.02539}, 2022.

\bibitem{b7}
T. Miyato, T. Kataoka, M. Koyama, and Y. Yoshida,
``Spectral normalization for generative adversarial networks,''
in \textit{Proc. Int. Conf. Learn. Represent. (ICLR)}, 2018.

\bibitem{b8}
A. Brock, J. Donahue, and K. Simonyan,
``Large scale GAN training for high fidelity natural image synthesis,''
in \textit{Proc. Int. Conf. Learn. Represent. (ICLR)}, 2019.

\bibitem{b9}
N. Srivastava, G. Hinton, A. Krizhevsky, I. Sutskever, and
R. Salakhutdinov,
``Dropout: A simple way to prevent neural networks from overfitting,''
\textit{J. Mach. Learn. Res.}, vol.~15, pp.~1929--1958, 2014.

\bibitem{b10}
A. Krogh and J. Hertz,
``A simple weight decay can improve generalization,''
in \textit{Adv. Neural Inf. Process. Syst. (NeurIPS)}, 1991.

\bibitem{b11}
A. Ross and F. Doshi-Velez,
``Improving the adversarial robustness and interpretability of deep
neural networks by regularizing their input gradients,''
in \textit{Proc. AAAI Conf. Artif. Intell.}, 2018.

\bibitem{b12}
K. Hornik, M. Stinchcombe, and H. White,
``Multilayer feedforward networks are universal approximators,''
\textit{Neural Netw.}, vol.~2, no.~5, pp.~359--366, 1989.

\bibitem{b13}
P. Ramachandran, B. Zoph, and Q. V. Le,
``Searching for activation functions,''
\textit{arXiv preprint arXiv:1710.05941}, 2017.

\bibitem{b14}
D. Hendrycks and K. Gimpel,
``Gaussian error linear units (GELUs),''
\textit{arXiv preprint arXiv:1606.08415}, 2016.

\bibitem{b15}
A. Vaswani, N. Shazeer, N. Parmar, J. Uszkoreit, L. Jones,
A. N. Gomez, L. Kaiser, and I. Polosukhin,
``Attention is all you need,''
in \textit{Adv. Neural Inf. Process. Syst. (NeurIPS)}, 2017.

\bibitem{b16}
A. Dosovitskiy et al.,
``An image is worth 16$\times$16 words: Transformers for image
recognition at scale,''
in \textit{Proc. Int. Conf. Learn. Represent. (ICLR)}, 2021.

\bibitem{b17}
P. Bartlett and S. Mendelson,
``Rademacher and Gaussian complexities: Risk bounds and structural
results,''
\textit{J. Mach. Learn. Res.}, vol.~3, pp.~463--482, 2002.

\bibitem{b18}
C. Zhang, S. Bengio, M. Hardt, B. Recht, and O. Vinyals,
``Understanding deep learning (still) requires rethinking
generalization,''
\textit{Commun. ACM}, vol.~64, no.~3, pp.~107--115, 2021.


\bibitem{b20}
J. Sokolić, R. Giryes, G. Sapiro, and M. R. D. Rodrigues,
``Robust large margin deep neural networks,''
\textit{IEEE Trans. Signal Process.}, vol.~65, no.~16,
pp.~4265--4280, 2017.

\bibitem{b21}
H. Zhang, Y. N. Dauphin, and T. Ma,
``Fixup initialization: Residual learning without normalization,''
in \textit{Proc. Int. Conf. Learn. Represent. (ICLR)}, 2019.
\bibitem{bouthillier2021}
X. Bouthillier, P. Delaunay, M. Bronzi, A. Trofimov, B. Nichyporuk, 
J. Szeto, N. Mohammadi Sepahvand, E. Raff, K. Madan, V. Voleti, 
S. E. Kahou, V. Michalski, T. Arbel, C. Pal, G. Varoquaux, and 
P. Vincent, ``Accounting for variance in machine learning benchmarks,'' 
in \emph{Proc. Mach. Learn. Syst. (MLSys)}, 2021.


\end{thebibliography}
\end{document}